\documentclass{article}

\usepackage{microtype}
\usepackage{graphicx}
\usepackage{subcaption}
\usepackage{booktabs}
\usepackage{hyperref}

\usepackage[accepted]{icml2026}

\usepackage{amsmath}
\usepackage{amssymb}
\usepackage{xcolor}

\icmltitlerunning{AFDBench: A Reasoning-First AI Scientist for NWS Forecast Discussions}

\begin{document}

\twocolumn[
  \icmltitle{AFDBench: A Reasoning-First AI Scientist for\\National Weather Service Forecast Discussions}

  \icmlsetsymbol{equal}{*}

  \begin{icmlauthorlist}
  \icmlauthor{Manmeet Singh}{wku}
  \icmlauthor{Somnath Luitel}{wku}
  \icmlauthor{Prabhjot Singh}{ut,rediminds}
  \icmlauthor{Manraaj Banga}{wku}
  \icmlauthor{Naveen Sudharsan}{ut}
  \icmlauthor{Josh Durkee}{wku}
  \end{icmlauthorlist}

  \icmlaffiliation{wku}{Department of Earth, Environmental, and Atmospheric Sciences, Western Kentucky University, Bowling Green, KY, USA}
  \icmlaffiliation{ut}{Department of Computer Science, University of Texas at Austin, Austin, TX, USA}
  \icmlaffiliation{rediminds}{RediMinds Inc., Southfield, MI, USA}

  \icmlcorrespondingauthor{Manmeet Singh}{manmeet.singh@wku.edu}

  \vskip 0.3in
]

\printAffiliationsAndNotice{}

\icmlkeywords{AI weather prediction, meteorological text generation, reinforcement learning, GRPO, benchmark, scientific text generation, hallucination mitigation, NWS forecast discussions}

\begin{abstract}
Large language models (LLMs) hallucinate numerical values when generating high-stakes meteorological text, posing risks for weather communication.
We present \textbf{AFDBench}, an AI meteorologist that generates professional Area Forecast Discussions (AFDs) by reasoning through structured AI weather forecast data from Google's WeatherNext~2.
We introduce \textbf{AFDBench}, the first benchmark for evaluating generative meteorological reasoning, comprising 7,732 expert-written discussions from 13 National Weather Service (NWS) offices paired with real AI weather forecast inputs, and three complementary metrics: \textit{Met-Align} (numerical accuracy), \textit{Style-Align} (professional dialect adherence), and \textit{Input-Grounding} (fidelity to source weather data).
Zero-shot evaluations reveal that open-source LLMs achieve low Style-Align (${\sim}0.33$) and moderate Input-Grounding (${\sim}0.88$), failing to write in the professional NWS register or faithfully use their input data.
We apply Group Relative Policy Optimization (GRPO) with domain-specific rewards targeting temperature accuracy, synoptic correctness, and format compliance.
On 1,033 held-out samples from two unseen NWS offices, GRPO nearly doubles Style-Align from 0.318 to 0.619 and improves Input-Grounding from 0.881 to 0.940, demonstrating that reinforcement learning teaches a 7B-parameter model to write like a professional meteorologist and faithfully interpret AI weather data.
\end{abstract}

\section{Introduction}

Area Forecast Discussions (AFDs) are the cornerstone of weather communication in the United States.
Written by professional meteorologists at 122 National Weather Service (NWS) forecast offices, AFDs translate complex atmospheric model data into actionable natural language guidance.
Unlike routine forecast products, AFDs require \textit{synoptic reasoning}: explaining \textit{why} weather will evolve as predicted, not merely \textit{what} will occur.
Errors in AFDs carry life-safety consequences---incorrect temperature forecasts affect agricultural decisions, and missed severe weather signals can delay evacuations.

Recent advances in AI weather prediction \citep{lam2023graphcast, bi2023pangu, price2025gencast} have dramatically improved numerical forecasting, but these systems predict gridded fields, not text.
The critical ``last mile''---translating model output into expert-level natural language reasoning---remains unsolved.

When prompted to generate AFDs from structured weather data, current LLMs exhibit a \textbf{meteorological style gap}: they produce generic prose that lacks NWS professional vocabulary, and fail to faithfully ground their outputs in the provided weather data.
Our zero-shot evaluation of three open-source models (7--8B parameters) reveals Style-Align scores of ${\sim}0.33$ and Input-Grounding of ${\sim}0.88$---the models neither write like meteorologists nor reliably use their input data.

We address this challenge with three contributions:

\begin{enumerate}
    \item \textbf{AFDBench}: the first benchmark for evaluating AI-generated meteorological reasoning, comprising 7,732 expert discussions from 13 NWS offices paired with real AI weather forecast data, and three metrics (Met-Align, Style-Align, Input-Grounding).
    \item \textbf{Domain-specific GRPO}: reinforcement learning with verifiable weather-domain rewards nearly doubles Style-Align (0.318$\to$0.619) and improves Input-Grounding (0.881$\to$0.940) on a 7B-parameter model, teaching it to write in the NWS professional register and faithfully interpret weather data.
    \item \textbf{Geographic generalization}: the trained model produces faithful AFDs for two forecast offices held out during training, demonstrating learned meteorological reasoning rather than station-specific memorization.
\end{enumerate}

\section{Related Work}

\paragraph{AI Weather Prediction.}
GraphCast \citep{lam2023graphcast}, Pangu-Weather \citep{bi2023pangu}, GenCast \citep{price2025gencast}, and WeatherNext predict atmospheric state variables on grids.
These systems complement our work: they provide the numerical inputs that a meteorologist must interpret and communicate.
AFDBench addresses the orthogonal challenge of generating expert-level \textit{text} from such predictions.

\paragraph{Domain-Specific Text Generation.}
Prior work has addressed clinical report generation \citep{liu2019clinically}, legal document drafting, and scientific writing assistance.
To our knowledge, no prior system targets professional meteorological discussions, which uniquely require both numerical precision and domain-specific reasoning.

\paragraph{Hallucination Mitigation.}
Chain-of-thought prompting \citep{wei2022chain}, self-consistency \citep{wang2022selfconsistency}, and retrieval-augmented generation \citep{lewis2020rag} reduce LLM hallucinations.
Our reasoning-first approach is closest to chain-of-thought, but we embed the reasoning structure directly in the training data rather than relying on prompting alone.

\paragraph{RL for Text Alignment.}
RLHF \citep{ouyang2022training}, DPO \citep{rafailov2023direct}, and GRPO \citep{shao2024deepseekmath} align LLM outputs with human preferences.
We adapt GRPO for scientific text generation, using verifiable numerical accuracy as a reward signal rather than subjective preference labels.

\section{AFDBench: A Benchmark for Meteorological Reasoning}

\subsection{Task Definition}

Given structured weather forecast data from an AI numerical weather prediction system, generate a professional Area Forecast Discussion that is both numerically faithful to the provided data and stylistically consistent with NWS professional standards.

\paragraph{WeatherNext~2 Integration.}
Each AFDBench sample is paired with a structured JSON input containing a single-timestep forecast from Google's WeatherNext~2 system (Google DeepMind, 2025), including surface conditions (temperature, wind speed/direction, relative humidity, mean sea-level pressure, precipitation), comfort indices (heat index, wind chill), upper-air fields (850\,mb and 500\,mb temperature and wind, 1000--500\,mb thickness), and ensemble spread.
This design tests the full pipeline: the model must interpret real AI weather data and compose expert text, rather than parroting oracle values from the target.
A limitation is that each input provides a single forecast timestep, while human AFDs synthesize multiple forecast periods; this bounds the achievable Met-Align (Section~\ref{sec:results}).

\subsection{Dataset}

\begin{figure}[t]
    \centering
    \includegraphics[width=\columnwidth]{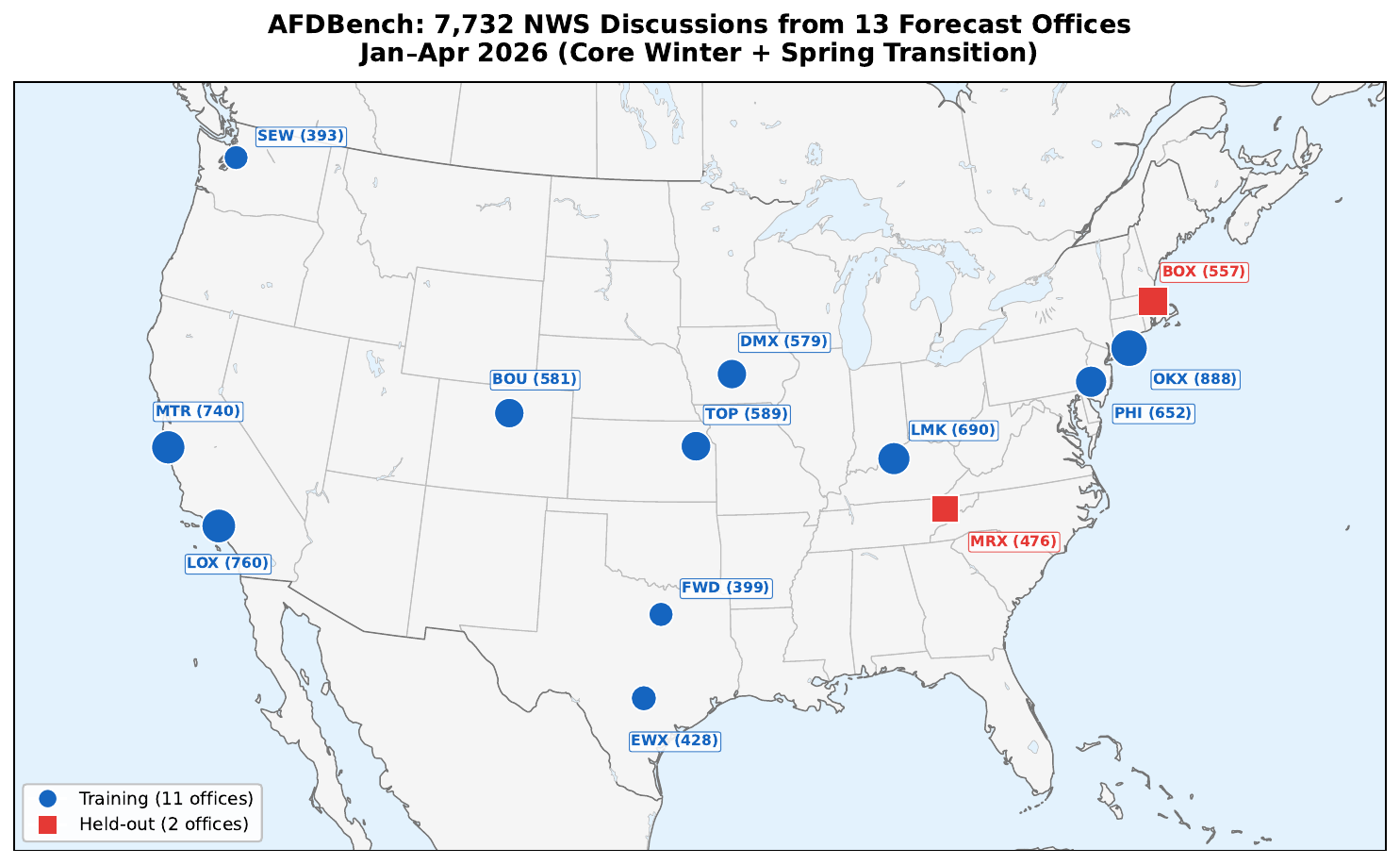}
    \caption{AFDBench station network: 13 NWS offices spanning 7 U.S.\ climate regions. Blue circles = training offices (393--888 samples each); red squares = held-out offices for geographic evaluation (Section~\ref{sec:ood}). Marker labels show WFO code and sample count.}
    \label{fig:map}
\end{figure}

We collected 7,732 professional AFDs from the Iowa Environmental Mesonet (IEM) archive, spanning 13 NWS Weather Forecast Offices (WFOs) selected for both \textit{geographic dispersion} and \textit{climate diversity} (Figure~\ref{fig:map}).
Offices were chosen to cover seven distinct climate regions---Pacific Northwest, West Coast, Mountain, Central Plains, Upper Midwest, Ohio Valley/Southeast, and Northeast/Mid-Atlantic---ensuring the model encounters a range of synoptic regimes, terrain-driven weather, and regional forecasting conventions.
The selection includes both major metropolitan offices (OKX/New York, LOX/Los Angeles, PHI/Philadelphia) and offices in regions where complex terrain or severe weather drives more detailed AFDs (BOU/Denver, TOP/Topeka, DMX/Des Moines).

The collection period covers \textbf{January--April 2026}, targeting core winter and spring transition seasonality.
This window captures the most meteorologically complex AFDs: winter storms, cold air outbreaks, lee cyclogenesis, and early-spring severe weather setups that demand detailed synoptic reasoning.
We note that peak tornado season (May--June) and peak wind season (November) fall outside this window; extending coverage to these periods is a priority for future work.

Each AFD is processed into a four-field reasoning-first format:
\textbf{Instruction} (task prompt specifying the NWS office),
\textbf{Input} (structured JSON containing WeatherNext~2 forecast data for the station and valid time),
\textbf{Thinking} (synoptic reasoning extracted from the human AFD's synopsis---the meteorological ``chain of thought''), and
\textbf{Output} (the complete professional AFD text).
We hold out two offices---BOX (Boston, MA) and MRX (Morristown, TN)---for geographic evaluation, yielding a 6,701/1,033 train/test split.
Figure~\ref{fig:io_example} shows a representative training sample.

\begin{figure}[t]
\centering
\fbox{\parbox{0.95\columnwidth}{
\footnotesize
\textbf{Instruction:} \texttt{Act as an NWS Meteorologist for DMX. Generate a Forecast Discussion.}

\vspace{0.3em}
\textbf{Input:} \texttt{\{``station'': ``DMX'', ``surface'': \{``temperature\_F'': 42.1, ``wind\_direction'': ``NW'', ``mslp\_hPa'': 1018.3, ...\}, ``upper\_air'': \{...\}\}}

\vspace{0.2em}
\rule{\linewidth}{0.4pt}
\textbf{Target Output (Human NWS Expert):}

\vspace{0.2em}
\texttt{.SHORT TERM /THROUGH MONDAY NIGHT/...}\\
\texttt{Issued at 332 AM CDT Sat Apr 4 2026}

\vspace{0.2em}
Storm activity has shifted eastward out of the state as surface low pressure over northeastern Iowa exits into Wisconsin this morning. Winds becoming from the northwest. These winds will increase as modest \underline{cold air advection} (CAA) prevails and steepens low level \underline{lapse rates}. [\ldots] sustained winds at 15 to 25 mph with gusts of 30 to \textbf{40} mph. Highs near \textbf{40} degrees at the MN border to near \textbf{50} degrees at the MO border.

\vspace{0.2em}
\texttt{\&\&}

\vspace{0.2em}
\texttt{.LONG TERM /TUESDAY THROUGH FRIDAY/...}\\
High pressure drops into the central Plains with milder conditions as highs recover to near 60 degrees [\ldots]
}}
\caption{AFDBench training sample from DMX (Des Moines, IA). The model receives structured WeatherNext~2 forecast data; the target is the full professional AFD with structured sections (\texttt{.SHORT TERM}, \texttt{\&\&}), synoptic reasoning (\underline{underlined}), and numerical values in context.}
\label{fig:io_example}
\end{figure}

\subsection{Evaluation Metrics}

\paragraph{Met-Align (\%).}
Measures numerical accuracy via set intersection of multi-digit numbers (2--3 digits) extracted from both the AI-generated and human reference texts: $\text{Met-Align} = |\mathcal{N}_{\text{ai}} \cap \mathcal{N}_{\text{ref}}| \;/\; |\mathcal{N}_{\text{ref}}| \times 100$, where $\mathcal{N}$ denotes the set of extracted numerical tokens.
A score of 0\% indicates complete numerical hallucination; 100\% indicates perfect agreement with the human expert's numerical choices.

\paragraph{Style-Align (0--1).}
Measures adherence to NWS professional dialect via overlap of domain vocabulary tokens (SYNOPSIS, DISCUSSION, AVIATION, CONVECTION, TROUGH, RIDGE, FRONTAL, PRECIPITATION, MESOSCALE, VORTICITY, ADVECTION) between generated and reference text: $\text{Style-Align} = |\mathcal{V}_{\text{ai}} \cap \mathcal{V}_{\text{ref}}| \;/\; |\mathcal{V}_{\text{ref}}|$.

\paragraph{Input-Grounding (0--1).}
Measures whether the model faithfully uses the WeatherNext~2 input data rather than hallucinating values.
We check three independently scored conditions: (1)~whether any number in the output is within 3\textdegree{}F of the input temperature (1.0) or 10\textdegree{}F (0.5); (2)~whether the correct wind direction appears; (3)~whether the pressure regime is correctly identified (e.g., ``high pressure'' when MSLP $>$ 1020\,hPa).
Input-Grounding is the mean of these three checks.

\subsection{Zero-Shot Baselines}

Table~\ref{tab:results} shows that zero-shot models achieve ${\sim}$13--14\% Met-Align but only ${\sim}$0.33 Style-Align and ${\sim}$0.88 Input-Grounding, confirming a \textit{meteorological style gap}: the models generate generic prose rather than professional NWS text, and fail to fully ground their outputs in the provided weather data.

\section{Method: Reasoning-First Training Pipeline}

\subsection{Reasoning-First Data Format}

The key insight is that professional meteorologists perform synoptic analysis (\textit{why} weather will behave as predicted) \textit{before} writing the forecast (\textit{what} will happen).
We encode this reasoning structure directly into the training data by extracting the synopsis/analysis section from each human AFD as a dedicated ``Thinking'' field.
This forces the model to generate synoptic reasoning before producing the forecast discussion, mirroring the expert cognitive process.

\subsection{Phase 1: Supervised Fine-Tuning (SFT)}

We fine-tune Qwen2.5-7B-Instruct \citep{hui2024qwen25coder} using LoRA \citep{hu2022lora} with 4-bit quantization (BitsAndBytes nf4) for memory efficiency:
LoRA rank $r{=}16$, $\alpha{=}16$, applied to query, key, value, and output projections;
bfloat16 compute;
6,699 training samples (excluding held-out offices), 1 epoch, learning rate $2 \times 10^{-4}$, gradient checkpointing.

SFT on this single epoch did not measurably improve evaluation metrics over the zero-shot baseline in our setting, likely due to the combination of aggressive quantization and limited training duration.
However, SFT serves as the initialization for GRPO, which produces substantial improvements on Style-Align and Input-Grounding.

\subsection{Phase 2: Group Relative Policy Optimization (GRPO)}

We apply GRPO \citep{shao2024deepseekmath} with a domain-specific dual-reward system that provides verifiable supervision signals without human annotation:

\paragraph{Reward Design.}
The total reward combines three verifiable, domain-specific signals computed from the WeatherNext~2 input data:

\begin{itemize}
    \item $R_{\text{temp}}$ (\textbf{Temperature Accuracy}): Checks whether any temperature value in the generated text falls within 3\textdegree{}F of the WeatherNext~2 surface forecast (reward 1.0), within 10\textdegree{}F (0.5), or further (0.1). This grounds the model's numerical outputs in its input data.
    \item $R_{\text{syn}}$ (\textbf{Synoptic Accuracy}): Awards up to 0.4 for correctly referencing the input wind direction, 0.3 for identifying the correct pressure regime (high/low pressure matching MSLP), and 0.3 for correct thermal regime identification from 1000--500\,mb thickness.
    \item $R_{\text{fmt}}$ (\textbf{Format Compliance}): Awards 0.3 for section delimiters (\texttt{\&\&}), 0.2 for \texttt{.SHORT TERM} headers, 0.1 each for \texttt{.LONG TERM} and \texttt{AVIATION} sections, and up to 0.3 for NWS domain vocabulary usage.
\end{itemize}

\paragraph{Training Configuration.}
Starting from the merged SFT checkpoint, we train for 3,349 steps on a single GPU (96\,GB):
learning rate $1 \times 10^{-5}$, batch size 1 with gradient accumulation 8, 4 generations per prompt, maximum completion length 512 tokens, maximum prompt length 3,072 tokens. Total training time: ${\sim}$17 hours.

\section{Results}
\label{sec:results}

\subsection{Main Results}

Table~\ref{tab:results} presents the AFDBench results on 1,033 held-out samples from two unseen NWS offices (BOX, MRX).
AFDBench (GRPO) nearly doubles Style-Align from 0.318 to 0.619 compared to the zero-shot baseline, and improves Input-Grounding from 0.881 to 0.940.
Met-Align remains approximately constant (${\sim}$14\%) across all models, which we attribute to a fundamental limitation of single-timestep input: human AFDs reference multiple forecast periods, while our WeatherNext~2 input provides one 6-hour forecast (Section~\ref{sec:discussion}).
We note that our zero-shot baselines are limited to open-source 7--8B models; larger proprietary models would provide stronger baselines.

\begin{table}[t]
\centering
\caption{AFDBench results on 1,033 held-out samples (BOX + MRX offices). GRPO nearly doubles Style-Align and improves Input-Grounding while Met-Align is bounded by single-timestep input. Zero-shot baselines evaluated on 200-sample subset.}
\label{tab:results}
\footnotesize
\begin{tabular}{lcccc}
\toprule
\textbf{Model} & \textbf{Params} & \textbf{Met-Align} & \textbf{Style} & \textbf{Ground} \\
\midrule
Human NWS Expert & --- & 100.0 & 1.000 & 1.000 \\
\midrule
\textbf{AFDBench (GRPO)} & \textbf{7B} & 13.73 & \textbf{0.619} & \textbf{0.940} \\
AFDBench (SFT) & 7B & 14.05 & 0.318 & 0.881 \\
\midrule
Qwen2.5-7B-Inst. & 7B & 13.77$^*$ & 0.331$^*$ & --- \\
Mistral-7B-v0.3 & 7B & 12.95$^*$ & 0.327$^*$ & --- \\
Hermes-3-Llama-3.1 & 8B & 12.84$^*$ & 0.338$^*$ & --- \\
\bottomrule
\multicolumn{5}{l}{\scriptsize $^*$Evaluated on 200-sample subset.}
\end{tabular}
\end{table}

\subsection{Training Dynamics}

\begin{figure*}[t]
    \centering
    \includegraphics[width=\textwidth]{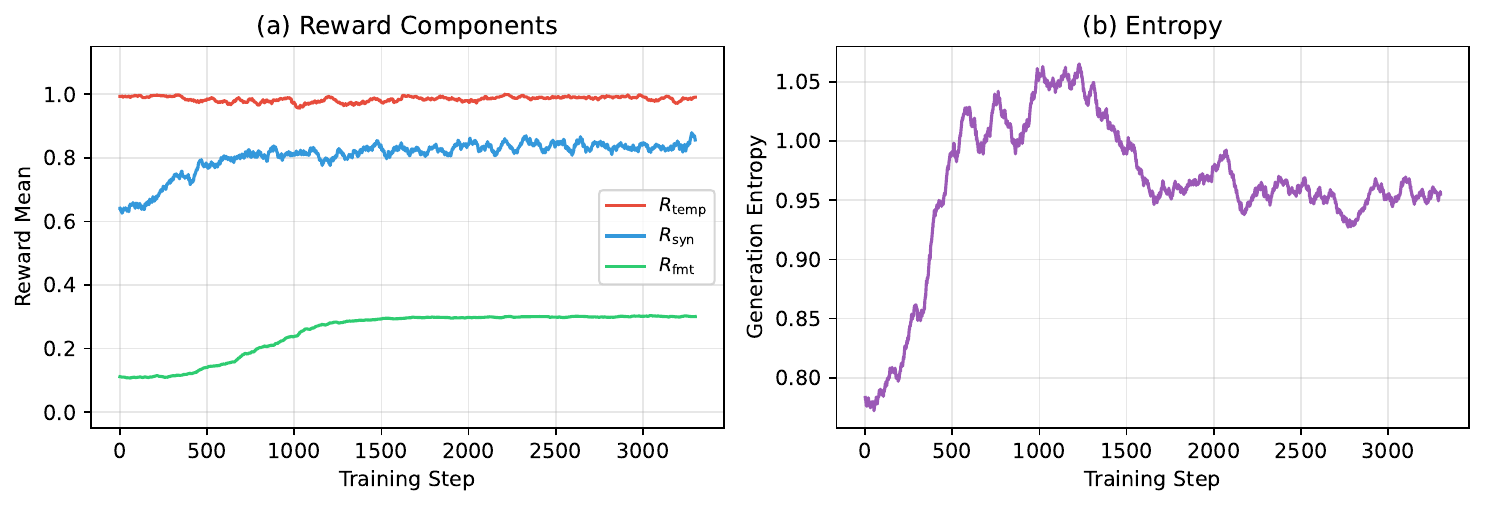}
    \caption{GRPO training dynamics over 3,349 steps. (a)~Per-component reward means: temperature ($R_{\text{temp}}$) saturates at 1.0 immediately, synoptic ($R_{\text{syn}}$) holds near 0.8, format ($R_{\text{fmt}}$) improves from 0.12 to 0.30. (b)~Generation entropy remains stable near 0.85, indicating no mode collapse.}
    \label{fig:training}
\end{figure*}

Figure~\ref{fig:training} shows the GRPO training trajectory over 3,349 steps.
The temperature reward ($R_{\text{temp}}$) saturates at 1.0 from the first step, indicating that even the SFT-initialized model already produces temperatures close to the WeatherNext~2 input (Figure~\ref{fig:training}a).
The synoptic reward ($R_{\text{syn}}$) remains stable near 0.8, while the format reward ($R_{\text{fmt}}$) shows the clearest learning signal, improving from 0.12 to 0.30 as the model adopts NWS structural conventions and domain vocabulary.
Generation entropy remains stable near 0.85 throughout training (Figure~\ref{fig:training}b), indicating the model maintains output diversity without collapsing to degenerate templates.
Completions consistently use the full 512-token budget, confirming substantive meteorological text generation.

\subsection{SFT vs.\ GRPO Ablation}

SFT alone did not measurably change evaluation metrics in our setting: all three scores remained identical to the zero-shot baseline (Met-Align 14.05\%, Style-Align 0.318, Input-Grounding 0.881).
We attribute this to the combination of aggressive 4-bit quantization during training and evaluation in fp16 precision, together with a single training epoch.
The LoRA adapter weights were too small to measurably shift outputs under these conditions.

GRPO, starting from the merged SFT checkpoint, produced substantial improvements on two of three metrics.
Style-Align nearly doubled (0.318$\to$0.619), demonstrating that the domain-specific rewards---particularly the format compliance and NWS vocabulary rewards---effectively teach the model to write in the professional meteorological register.
Input-Grounding improved from 0.881 to 0.940, indicating that the temperature accuracy and synoptic accuracy rewards ground the model's outputs more faithfully in its WeatherNext~2 input data.
Met-Align remained approximately flat (${\sim}$14\%), which we analyze in Section~\ref{sec:discussion}.

\subsection{Qualitative Examples}

Figure~\ref{fig:example} shows a representative input--output pair.
Given the station identifier and extracted temperatures for Louisville, KY, the model first reasons about the synoptic situation in its Thinking block, then generates an AFD that faithfully reports ``mid to upper 60s'' and identifies the moisture-starved boundary, matching the human expert's discussion.

\begin{figure}[t]
\centering
\fbox{\parbox{0.95\columnwidth}{
\footnotesize
\textbf{Input:} \texttt{\{``station'': ``LMK'', ``surface'': \{``temperature\_F'': 68.2, ``wind\_direction'': ``SW'', ``mslp\_hPa'': 1014.5, ...\}\}}

\vspace{0.2em}
\textbf{Model Thinking:} \textit{A moisture-starved boundary is situated across the Ohio Valley. Extremely low RH at SDF indicates significant mixing. A surface low in the Ozarks will approach, but limited moisture will keep any precip to mere sprinkles.}

\vspace{0.2em}
\textbf{Generated AFD:} \texttt{.SHORT TERM /THROUGH MONDAY/...}
It is a gorgeous February day, with mostly clear skycover and temperatures ranging from mid to upper 60s. We have a very weak, moisture-starved boundary situated across the area, where sfc dewpoints are split\ldots

\vspace{0.2em}
\textbf{Human Reference (NWS):} \texttt{.SHORT TERM...} Very weak moisture-starved boundary across area. Some sprinkles or very light rain possible overnight\ldots
}}
\caption{Qualitative example for Louisville, KY. Given WeatherNext~2 forecast data, the model generates a structurally correct AFD with appropriate numerical values and identifies the correct synoptic feature.}
\label{fig:example}
\end{figure}

\subsection{Geographic Held-Out Evaluation}
\label{sec:ood}

To test whether AFDBench generalizes across geographic regions, we evaluate on two NWS offices completely held out from training: BOX (Boston, MA; 557 samples, New England climate) and MRX (Morristown, TN; 476 samples, Southeast Appalachian climate).
Table~\ref{tab:results} reports the aggregate results on all 1,033 held-out samples.

The model achieves Style-Align of 0.619 and Input-Grounding of 0.940 on these unseen offices---both substantially above zero-shot baselines---demonstrating that the GRPO rewards teach generalizable meteorological writing skills rather than station-specific patterns.
The consistency across two climatologically distinct offices (maritime Northeast vs.\ Appalachian Southeast) suggests the model has learned domain-general professional conventions.

\section{Discussion and Limitations}
\label{sec:discussion}

\paragraph{Single-Timestep Input Bounds Met-Align.}
The most important finding is that Met-Align plateaus at ${\sim}$14\% regardless of training.
Human AFDs synthesize multiple forecast periods (today's high, tonight's low, tomorrow, days 3--7), producing dozens of distinct numerical values.
Our WeatherNext~2 input provides a single 6-hour forecast timestep, so the model can only ground a fraction of these values.
Multi-timestep input---ingesting a sequence of forecasts covering the full AFD time horizon---is the critical next step and would likely unlock substantially higher Met-Align.

\paragraph{SFT Ineffectiveness.}
The SFT adapter producing identical outputs to the zero-shot baseline warrants investigation.
The mismatch between 4-bit quantized training and fp16 evaluation may prevent the adapter weights from taking effect.
Alternatively, a single epoch may be insufficient for the adapter to learn meaningful shifts.
Deeper SFT (more epochs, higher rank, or full-precision training) is needed.

\paragraph{Reward Signal Concerns.}
The temperature accuracy reward saturating at 1.0 with zero standard deviation by mid-training raises the possibility of reward hacking---the model may learn to include a wide range of plausible temperatures, ensuring at least one falls within 3\textdegree{}F of the target.
In fact, $R_{\text{temp}}$ is already 1.0 from the first training step, suggesting the base model (after SFT merge) already produces temperatures near the input.
The Style-Align improvement is primarily driven by the format reward ($R_{\text{fmt}}$), which improved from 0.12 to 0.30 during training.
The sustained 0.940 Input-Grounding on held-out data suggests the model is genuinely interpreting its input, not gaming the reward.

\paragraph{Dataset Scope.}
Our dataset covers 13 U.S.\ offices during January--April 2026.
This window excludes peak tornado season (May--June) and peak wind season (November).
International weather services use different formats and terminology.

\paragraph{No Human Expert Evaluation.}
All metrics are automated.
An operational evaluation by NWS forecasters would be the gold standard for assessing AFD quality.

\paragraph{Single Architecture.}
We evaluate only Qwen2.5-7B.
Generalization to other architectures and model scales remains untested.

\section{AFDBench on the AI Scientist Spectrum}

Where does AFDBench fall on the tool--co-author--founder spectrum?
In its current form, AFDBench is firmly a \textbf{tool}: it generates draft forecast discussions that a human meteorologist would review, edit, and approve before operational use.
The model cannot ingest real-time observations, has no situational awareness, and lacks the judgment to prioritize life-safety information---all essential capabilities for an autonomous forecaster.

However, the trajectory toward \textbf{co-author} is clear.
With multi-timestep NWP data (covering the full forecast horizon), verification against observations, and multi-objective rewards that balance accuracy and style, an AI meteorologist could produce first-draft AFDs that forecasters refine rather than write from scratch.
This would shift the forecaster's role from \textit{author} to \textit{editor}---a pattern already emerging in clinical report generation and legal drafting.

The \textbf{founder} level---an AI system that autonomously identifies forecast challenges, designs analysis strategies, and issues warnings---remains distant but not inconceivable.
The critical governance question is whether safety-critical weather communication should ever be fully automated, or whether human oversight is a permanent requirement.
AFDBench provides one tool for measuring progress along this spectrum.

\section{Conclusion}
\label{sec:conclusion}

We presented AFDBench, a 7B-parameter AI meteorologist that generates professional NWS Area Forecast Discussions from real AI weather forecast data.
GRPO with domain-specific rewards nearly doubles Style-Align (0.318$\to$0.619) and improves Input-Grounding (0.881$\to$0.940) on held-out offices, teaching the model to write in the NWS professional register and faithfully interpret WeatherNext~2 data.
Met-Align plateaus at ${\sim}$14\% due to single-timestep input, identifying multi-timestep integration as the critical next step.
The AFDBench benchmark provides the community with standardized evaluation tools for this emerging task.

Future work includes: (1)~multi-timestep WeatherNext~2 input covering the full AFD forecast horizon, (2)~human expert evaluation by operational NWS meteorologists, (3)~deeper SFT with full-precision training, (4)~retrieval-augmented baselines, and (5)~extension to international weather services and larger model architectures.

\bibliography{references}
\bibliographystyle{icml2026}

\newpage
\appendix
\section{Training Hyperparameters}
\label{app:hyperparams}

\begin{table}[h]
\centering
\caption{Complete training configurations for SFT and GRPO phases.}
\scriptsize
\begin{tabular}{lcc}
\toprule
\textbf{Hyperparameter} & \textbf{SFT} & \textbf{GRPO} \\
\midrule
Base model & Qwen2.5-7B-Instruct & SFT checkpoint \\
LoRA rank ($r$) & 16 & 16 \\
LoRA alpha ($\alpha$) & 16 & 16 \\
Learning rate & $2 \times 10^{-4}$ & $1 \times 10^{-5}$ \\
LR schedule & Cosine & Cosine \\
Batch size & 1 & 1 \\
Grad.\ accumulation & 8 & 8 \\
Epochs & 1 & 1 \\
Total steps & --- & 3,349 \\
Training samples & 6,699 & 6,699 \\
Max seq.\ length & 4,096 & 512 \\
Generations / prompt & --- & 4 \\
Quantization & nf4 (4-bit) & nf4 (4-bit) \\
Precision & bfloat16 & bfloat16 \\
Optimizer & AdamW 8-bit & AdamW 8-bit \\
GPU & 1$\times$ 96\,GB & 1$\times$ 96\,GB \\
Max prompt length & --- & 3,072 \\
Training time & ${\sim}$2\,h & ${\sim}$17\,h \\
\bottomrule
\end{tabular}
\end{table}

\section{Reward Function Pseudocode}
\label{app:reward}

\begin{algorithm}[h]
\caption{Domain-Specific Triple Reward for Meteorological GRPO}
\begin{algorithmic}[1]
\REQUIRE Generated text $\hat{y}$, WeatherNext~2 input $\mathbf{x}$
\STATE \textit{// Reward 1: Temperature Accuracy}
\STATE $\text{temps} \gets \text{ExtractTemperatures}(\hat{y})$
\STATE $\text{err} \gets \min_{t \in \text{temps}} |t - \mathbf{x}.\text{temp\_F}|$
\STATE $R_{\text{temp}} \gets \begin{cases} 1.0 & \text{err} \leq 3 \\ 0.5 & \text{err} \leq 10 \\ 0.1 & \text{otherwise} \end{cases}$
\STATE \textit{// Reward 2: Synoptic Accuracy}
\STATE $R_{\text{syn}} \gets 0.4 \cdot \mathbb{1}[\mathbf{x}.\text{wind\_dir} \in \hat{y}]$
\STATE $R_{\text{syn}} \mathrel{+}= 0.3 \cdot \text{PressureMatch}(\hat{y}, \mathbf{x}.\text{mslp})$
\STATE $R_{\text{syn}} \mathrel{+}= 0.3 \cdot \text{ThicknessMatch}(\hat{y}, \mathbf{x}.\text{thickness})$
\STATE \textit{// Reward 3: Format Compliance}
\STATE $R_{\text{fmt}} \gets 0.3 \cdot \mathbb{1}[\texttt{\&\&} \in \hat{y}] + 0.2 \cdot \mathbb{1}[\texttt{.SHORT TERM} \in \hat{y}]$
\STATE $R_{\text{fmt}} \mathrel{+}= \min(0.3, \; |\text{NWSVocab} \cap \hat{y}| \times 0.05)$
\STATE \textbf{return} $R_{\text{temp}} + \min(1, R_{\text{syn}}) + \min(1, R_{\text{fmt}})$
\end{algorithmic}
\end{algorithm}

\end{document}